\documentclass[]{article}
\usepackage{graphicx}
\usepackage{subcaption}
\usepackage{amsmath}
\usepackage{amssymb}
\usepackage{float}
\usepackage{url}
\usepackage{xcolor}
\usepackage{makecell}
\usepackage{cite}
\usepackage{tabularx}
\usepackage{paralist}
\usepackage{array}
\usepackage{booktabs}
\usepackage{gensymb}            
\usepackage{tikz}
\usepackage[margin=1in]{geometry}

\usepackage{caption}
\usepackage[utf8]{inputenc}
\usepackage{pgfplots}
\DeclareUnicodeCharacter{2212}{−}
\usepgfplotslibrary{groupplots,dateplot}
\usetikzlibrary{patterns,shapes.arrows}
\pgfplotsset{compat=newest}
\usepackage{pgf}

\newcommand{\eg}{\textit{e.g.}, }

\title{\LARGE \bf Using Monocular Vision and Human Body Priors\\ for AUVs to Autonomously Approach Divers *}

\author{Michael Fulton$^{1}$, Jungseok Hong$^{2}$, and Junaed Sattar$^{3}$
\thanks{The authors are with the Department of Computer Science and Engineering and the Minnesota Robotics Institute,
        University of Minnesota Twin Cities, Minneapolis, MN, USA.
        {\tt\small \{$^{1}$fulto081,$^{2}$jungseok,$^{3}$junaed\}@umn.edu}}%
\thanks{*This work was supported by the US National Science Foundation awards IIS-\#1637875 \& \#00074041, and the MnRI Seed Grant. The first two authors made equal contribution to the work and should both be cited as first author (ex. \textit{Fulton and Hong et al.})}%
}

\begin{document}

\maketitle
\thispagestyle{empty}
\pagestyle{empty}

\begin{abstract}
Direct communication between humans and autonomous underwater vehicles (AUVs) is a relatively underexplored area in human-robot interaction (HRI) research, although many tasks (\eg surveillance, inspection, and search-and-rescue) require close diver-robot collaboration. Many core functionalities in this domain are in need of further study to improve robotic capabilities for ease of interaction. One of these is the challenge of autonomous robots approaching and positioning themselves relative to divers to initiate and facilitate interactions. Suboptimal AUV positioning can lead to poor quality interaction and lead to excessive cognitive and physical load for divers. In this paper, we introduce a novel method for AUVs to autonomously navigate and achieve diver-relative positioning to begin interaction.
Our method is based only on monocular vision, requires no global localization, and is computationally efficient.
We present our algorithm along with an implementation of said algorithm on board both a simulated and physical AUV, performing extensive evaluations in the form of closed-water tests in a controlled pool.
Analysis of our results show that the proposed monocular vision-based algorithm performs reliably and efficiently operating entirely on-board the AUV.

\end{abstract}

\section{INTRODUCTION}
\label{sec:intro}
\begin{figure}[t]
    \centering
    \begin{subfigure}[b]{0.48\linewidth}
        \centering
        \includegraphics[width=\linewidth, trim=0cm 0cm 0cm 1.8cm, clip]{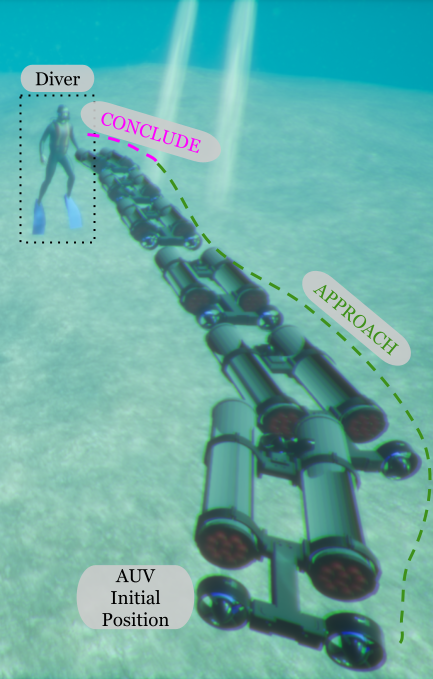}
        \vspace{-4mm}
        \label{fig:intro:a}
        \caption{Head-on}
    \end{subfigure}
    \begin{subfigure}[b]{0.48\linewidth}
    \centering
        \includegraphics[width=\linewidth]{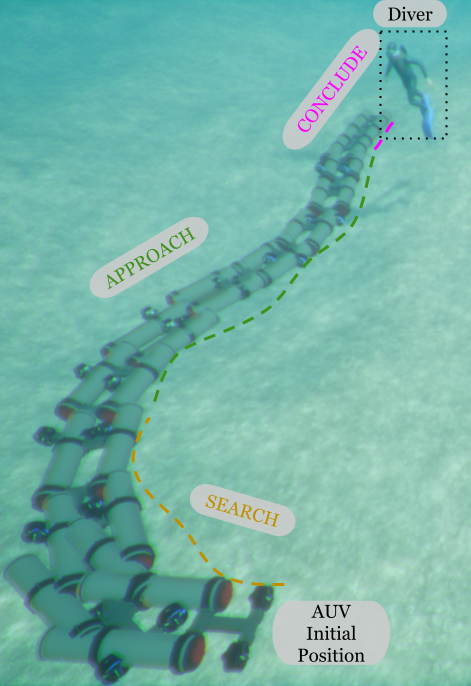}
        \vspace{-4mm}
        \label{fig:intro:b}
        \caption{Turned-away}
    \end{subfigure}
    \caption{Two examples of ADROC diver approaches: (a) a diver is in the AUV's field of view, so the AUV simply moves to the diver, (b) no diver is visible, so ADROC begins a search procedure, approaching once the diver is found. }
    \label{fig:intro}
    \vspace{-4mm}
\end{figure}

Autonomous underwater vehicles (AUV) have traditionally been used for standalone missions, with limited or no direct human involvement, in applications where it is infeasible for humans to closely collaborate with the robots (\eg long-term oceanographic surveys~\cite{bellingham_oceanographic_1996, hwang_adaptive_2019}).
However, in recent decades, the advent of smaller AUVs~\cite{dudek_aqua_2007, miskovic_caddyyear3_2017, loco_paper_2020} suitable for working closely with humans (termed co-AUVs) has enabled robots and humans to collaborate on tasks underwater.
Over the last few decades, a variety of human-robot interaction (HRI) capabilities have been developed for co-AUVs to facilitate this work: from gestural programming~\cite{islam_dynamicreconfigurationmission_2018, chavez_robustgesturebased_2018,islam_gestures_2019} and various modes of robot communication~\cite{demarco_underwaterhumanrobot_2014,fulton_robotcommunicationvia_2019} to diver following~\cite{sattar_tracking_2007,islam_2018_toward}.
Unfortunately, underwater HRI systems typically assume that both divers and AUVs are in the ideal position for interaction, an assumption that does not always hold true. 
When working in the challenging, unstructured underwater environment, human dive partners are often required to find one another to begin their interactions, and the same is true of AUVs working with divers.
Despite its critical importance, the problem of \textbf{how an AUV should find its dive partner and position itself to facilitate interaction} has yet not been addressed.
While divers could swim to the AUV themselves, placing the burden of maintaining an appropriate position for interaction on the divers both adds to their existing physical and cognitive loads and robs them of precious time and oxygen.
This increases the overhead of working with AUVs, decreasing their usefulness and their adoption rate in the field.
Creating the ability for an AUV to seek out and position itself relative to its operator (as seen in Fig. \ref{fig:intro}) will increase the utility of HRI-enabled co-AUVs in many underwater tasks.


The problem of an AUV approaching a diver to facilitate underwater human-robot interactions has not been sufficiently addressed in the past, despite the fact that robots approaching humans in indoor scenarios (offices, warehouses, etc.) has been studied for decades~\cite{dautenhahn_how_2006,syrdal_doing_2006}. 
Service robots, for example, have been approaching humans effectively for years, but the task of achieving this kind of behavior underwater is much more challenging due to the environmental challenges of the domain: \eg degraded sensing quality, less reliable localization and motion planning, and lack of wireless communication.
This follows the general trend of underwater HRI research: many problems which have been functionally solved for terrestrial robots have yet to be addressed as successfully underwater due to the increased challenges of the domain.
When we consider the realm of underwater HRI, the sub-field with the most relation to diver approach scenarios is the topic of diver following, which has been widely studied.
Diver following, however, is a very different problem.
In diver following scenarios, there typically are no strict constraints on the distance between diver and AUV, no precise evaluation of the position of the AUV relative to the diver, and no functionality for searching for a target who has been lost or is as of yet unseen.
All of these are critical aspects of the AUV diver approach scenario.
Furthermore, the few works which do impose strict constraints on the position and distance of an AUV relative to a diver require the use of expensive sensors (\eg stereo cameras, sonar, Doppler velocity log (DVL), ultra-short baseline (USBL)/long baseline (LBL) localization hardware).
This prevents AUVs without those sensors from taking advantage of the precise diver-relative navigation abilities of these works, abilities key to effective diver approach scenarios.

No existing diver-based AUV navigation method is capable of reliably approaching a diver (robustly achieving a stable, desired relative position, orientation, and distance) using only monocular visual information. 
The absence of such functionality greatly reduces the effectiveness of interactive AUVs, preventing them from reaching their full potential.
To address this need, we present Autonomous Diver-Relative Operator Configuration (ADROC): a novel algorithm that gives any AUV with a camera the ability to approach a diver.
ADROC provides navigation to an ideal location for interaction with a diver quickly and with a high rate of success.
We present the modular design of our algorithm and brief descriptions of the components currently in use.
We also present a cascaded method for estimating the distance to a diver, first using bounding box detections from a long range, then refining the estimate using body pose estimation and biological priors at closer ranges.
Following this, we discuss two pool experiments in which an AUV approached nine different target divers each from nine different initial positions using ADROC.
Our method achieved $132$ approaches successfully out of the $162$ attempted, for an overall success rate of \textbf{$81\%$} with an average approach time of $26$ seconds.
\\ \\
\noindent\textbf{Contributions:}
In this paper, we present the following:
\begin{itemize}
    \item Autonomous Diver-Relative Operator Configuration (ADROC), a novel algorithm for AUV navigation to a position ideal for interaction with a diver.
    \item A monocular-vision-based method for estimating approximate distance based on diver biological priors.
    \item Analysis of pool experimentation with ADROC.
      
\end{itemize}
\section{RELATED WORK}
\label{sec:related}

\begin{figure}
    \vspace{2mm}
	\centering
	 \begin{subfigure}[b]{0.45\textwidth}
        \includegraphics[width=\textwidth,trim={4cm 2cm 2cm 6.7cm},clip]{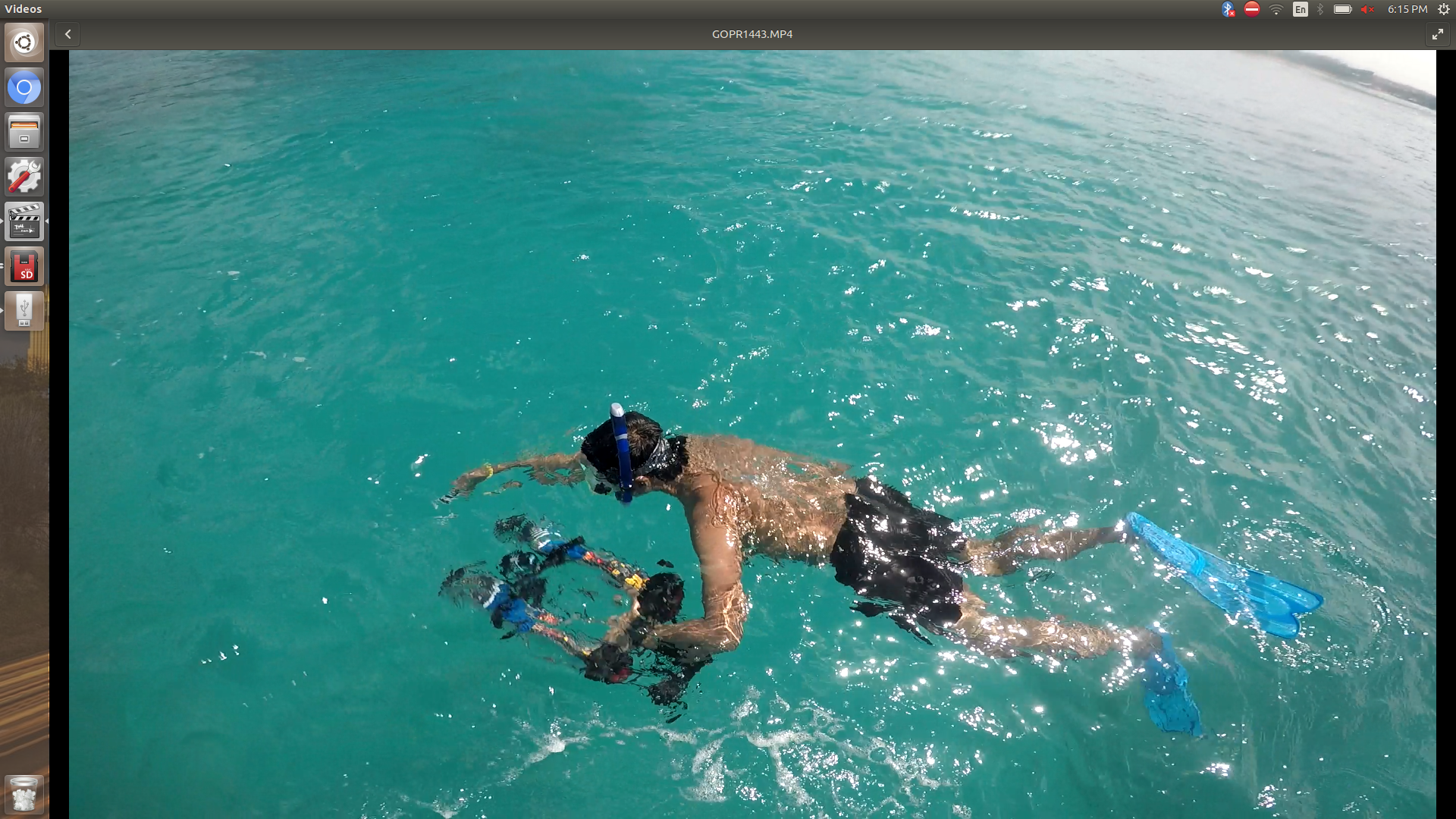}
    \end{subfigure}
    \begin{subfigure}[b]{0.45\textwidth}
        \includegraphics[width=\textwidth,trim={0cm 0cm 0cm 0.2cm},clip]{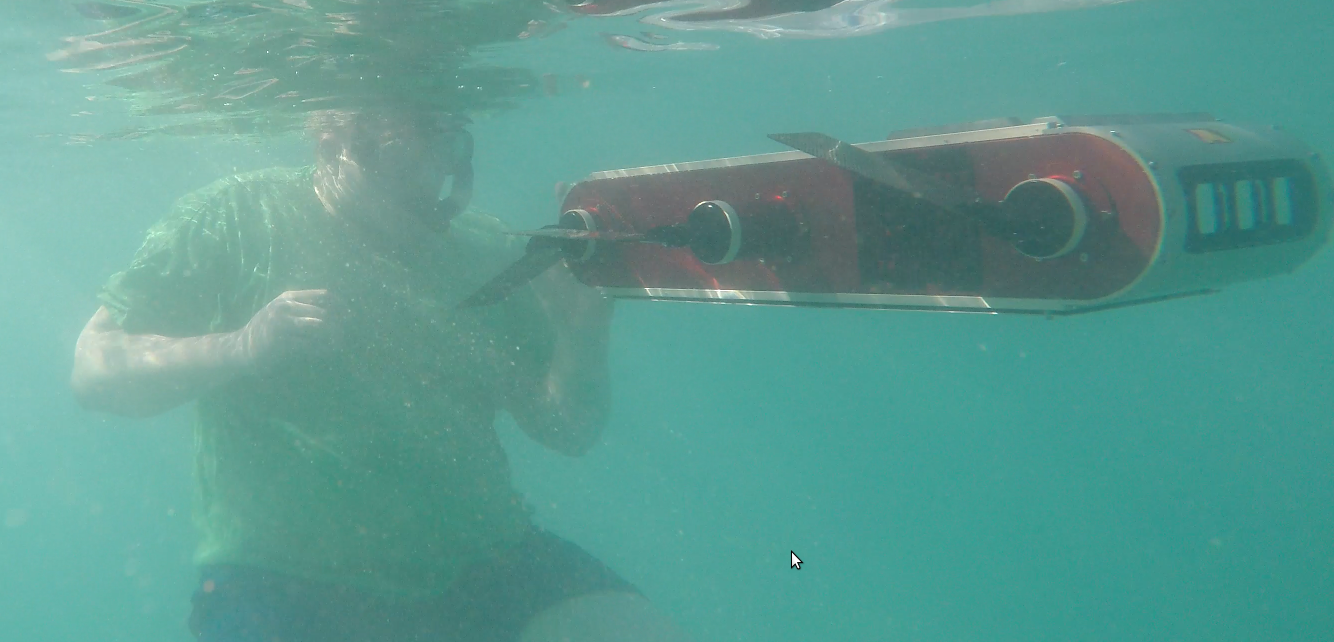}
    \end{subfigure}
    \begin{subfigure}[b]{0.45\textwidth}
        \includegraphics[width=\textwidth,trim={0cm 1.5cm 0cm 0cm},clip]{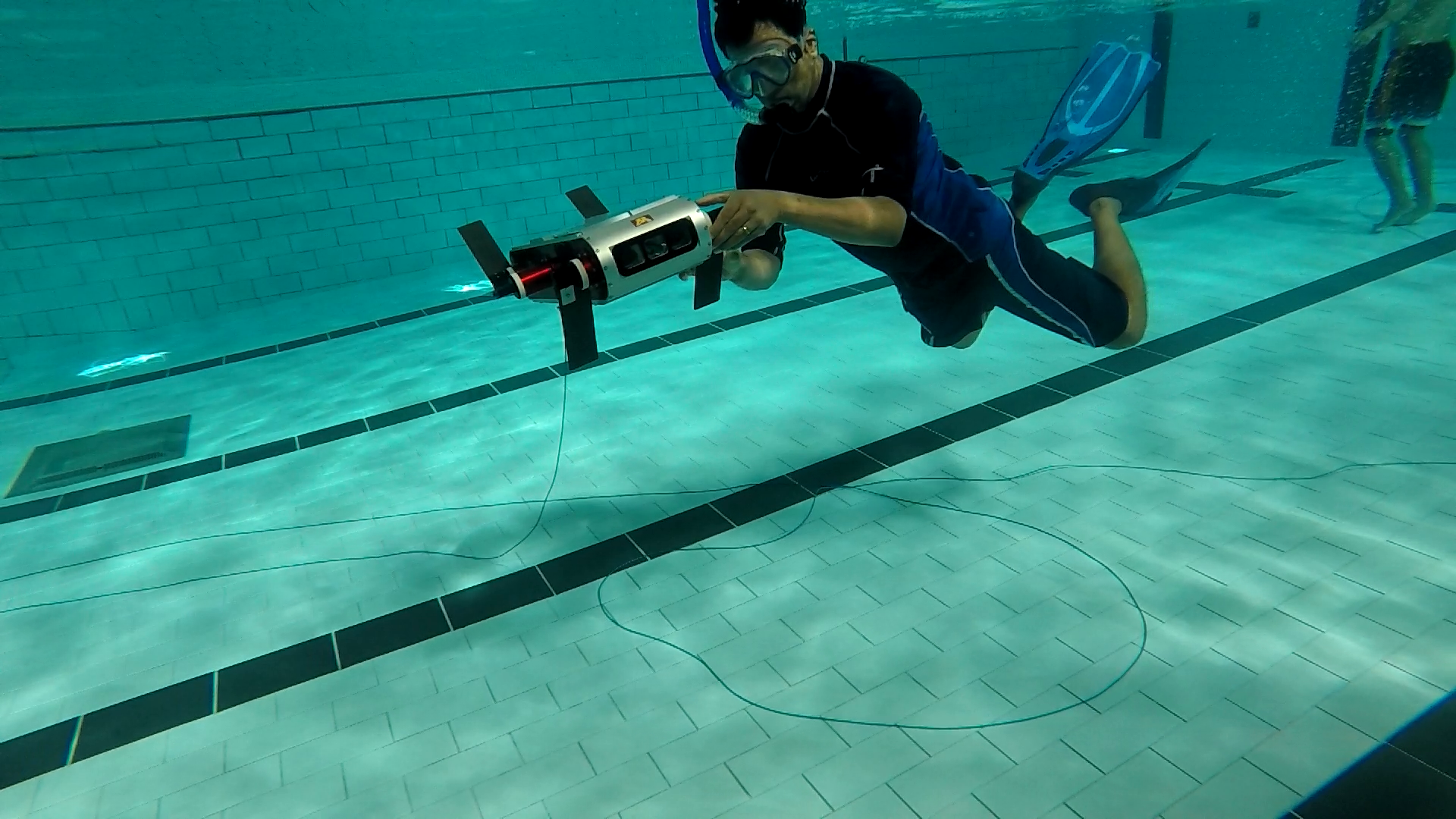}
    \end{subfigure}
    \begin{subfigure}[b]{0.45\textwidth}
        \includegraphics[width=\textwidth,trim={5cm 4.5cm 0cm 1.2cm},clip]{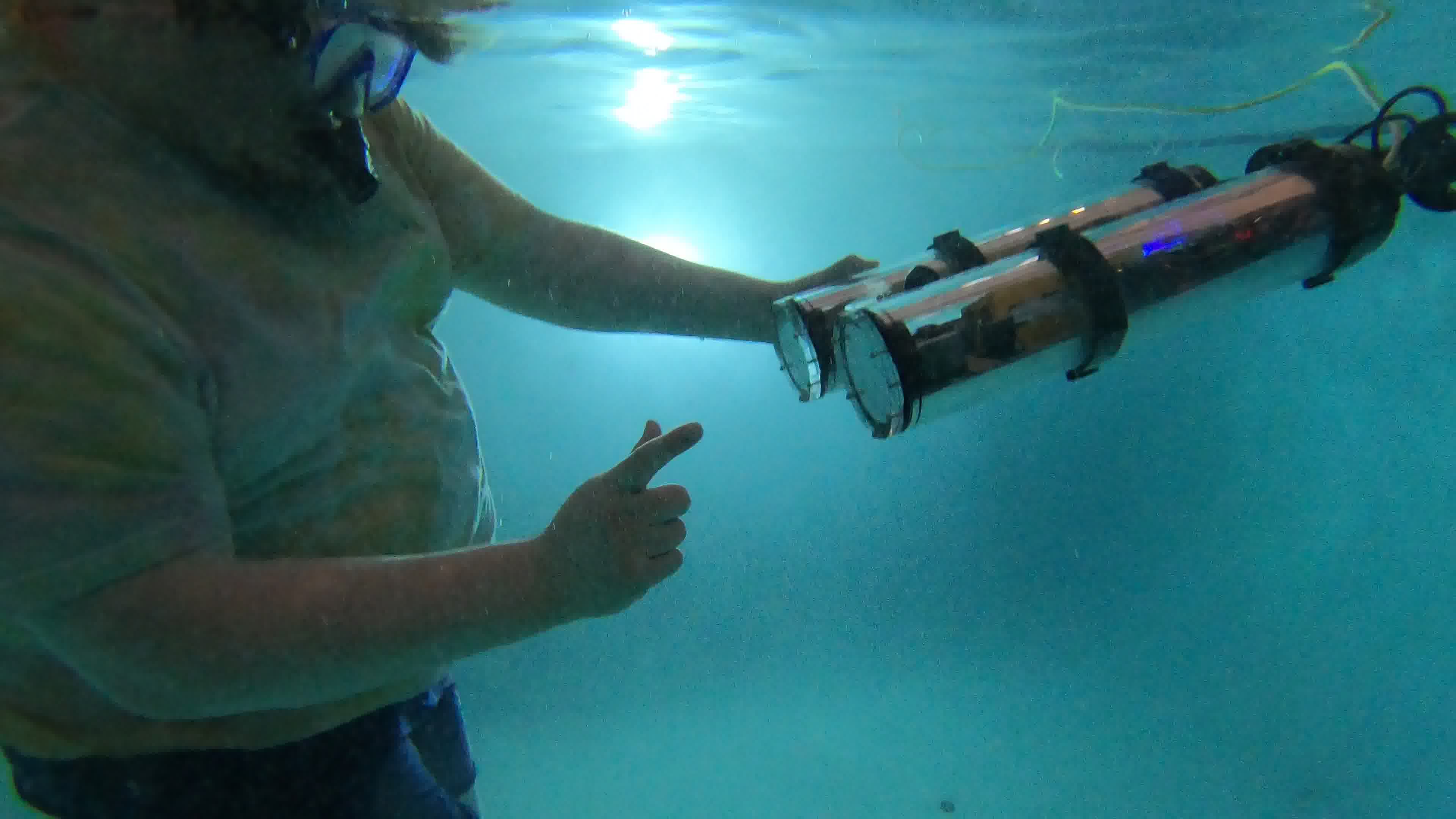}
    \end{subfigure}
    \caption{Robots and humans interacting underwater.}
    \label{fig:field_interaction}
    \vspace{-5mm}
\end{figure}


\subsection{Robots Approaching Humans}
\label{sec:related:approach}
The problem of robots approaching human targets has been extensively explored in HRI research, having been mostly considered in the context of service robots~\cite{brandl_serviceapproach_2016}.
A large amount of robot approach research focuses on the effect of the proxemic behaviors of human interactants: the way in which the interactant's concept of personal space affects how a robot should approach them for the most optimal interactions~\cite{dautenhahn_how_2006,syrdal_doing_2006,walters_human_2008,takayama_proxemics_2009,brandl_serviceapproach_2016}. 
Research on robot approaches in field environments (which pose more of a challenge in terms of sensing and navigation) is rare, but some research has explored approaches in aerial environments~\cite{monajjemi_cometome_2016}, focusing on the algorithms necessary to enable an unmanned aerial vehicle (UAV) to approach a human interactant.
However, there is no work on this topic for AUVs.
The majority of research related to approaching humans with an AUV is instead focused on the broader problem of diver-based AUV navigation.

\subsection{Diver-Based AUV Navigation}
\label{sec:related:diver-relative}
Diver-based AUV navigation is a common topic in AUV research for co-AUVs (AUVs designed to work collaboratively with humans) dealing with the varied scenarios in which an AUV must navigate its environment with respect to a diver.
The most common form of this problem is diver following~\cite{sattar_tracking_2007,islam_2018_toward,remmas_divertracking_2021}, a subset of the larger field of person-following research~\cite{islam_personfollowing_2019}.
Many diver following works in recent years have utilized deep neural networks to detect a diver, then used closed-loop control algorithms to follow the diver based on detections.
Some works have gone further than the simple scenario of diver following, such as the work of Nad et al.~\cite{nad_diver_tracking_2020}, which utilizes acoustic and visual information, allowing the robot to both act as a follower and leader.
However, there are fundamental differences between following a diver and navigating to a position relative to a diver to facilitate further interaction.
Approaching a diver requires complex algorithmic structures not typically found in diver following: the ability to actively search for lost or unseen divers, evaluation of the quality of relative position for interaction, and strict constraints on the distance between diver and robot.
While many diver following methods contain some of the same capabilities necessary for our task (detection of a diver and navigation based on that detection), diver following is a different problem with different goals.

 \begin{figure*}
    \centering
    \vspace{2mm}
    \includegraphics[width=\linewidth, trim=0cm 0cm 0cm .5cm, clip]{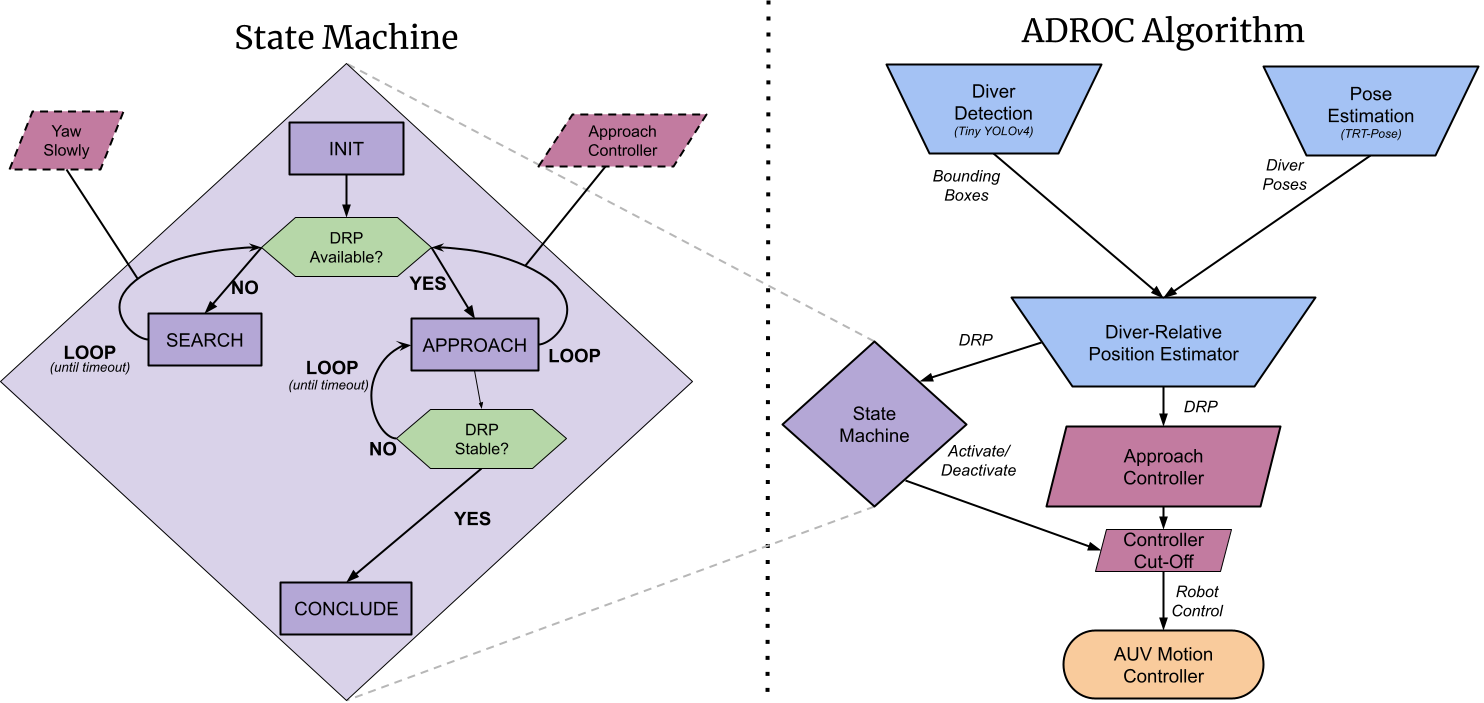}
    \caption{Diagram showing the ADROC algorithm \textit{(right)} with detail on the state machine \textit{(left)}. Different components of the algorithm (perception, states, approach controller, and conditions) are presented in consistent colors and shapes.}
    \label{fig:aoc_algorithm}
    \vspace{-4mm}
\end{figure*}

\subsection{Estimation of Distance to Diver}
\label{sec:related:distance}
A capability common to both our task (AUV approaching a diver) and other diver-based navigation tasks is the estimation of the relative position and range of a diver.
The majority of diver-following methods do not estimate the distance to a diver accurately.
Methods which do depend on the high fidelity data exploit sonar, stereo camera, USBL, and DVL sensors.
A method for distance estimation that avoids the most expensive sensors is the use of stereo vision~\cite{scharstein2002taxonomy}.
Stereo-based distance estimation could be useful to estimate the range to a diver at close range with a specific environment setup, but it will be prone to errors if not specifically tuned to its environment. 
This is because the camera parameters are affected when deployed in a new optical environment and this change could degrade the accuracy of the distance estimation. 
Additionally, the error in the depth estimate increases quadratically as the distance from the camera increases~\cite{michels2005high, gallup2008variable}, which limits the range of any algorithm depending on it. 
Our algorithm achieves reliable distance estimation using only low-cost monocular camera data, distinguishing it from the previously discussed work by improving distance estimation for diver-relative navigation without the use of complex and expensive sensors.

\section{AUTONOMOUS DIVER-RELATIVE\\ OPERATOR CONFIGURATION}
\label{sec:aoc}
To enable AUVs to track and approach divers for the purpose of initiating interactions, we present the Autonomous Diver-Relative Operator Configuration (ADROC) algorithm. 
The proposed algorithm was implemented using the LoCO AUV~\cite{loco_paper_2020}, but due to its minimal sensor requirements, it could be transferred to any AUV with a monocular camera and five or more degrees of freedom.

\begin{figure*}[t]
\vspace{2mm}
\centering
    \begin{subfigure}{.45\textwidth}
      \includegraphics[width=\linewidth, trim= 0cm 0.8cm 0cm 0cm, clip]{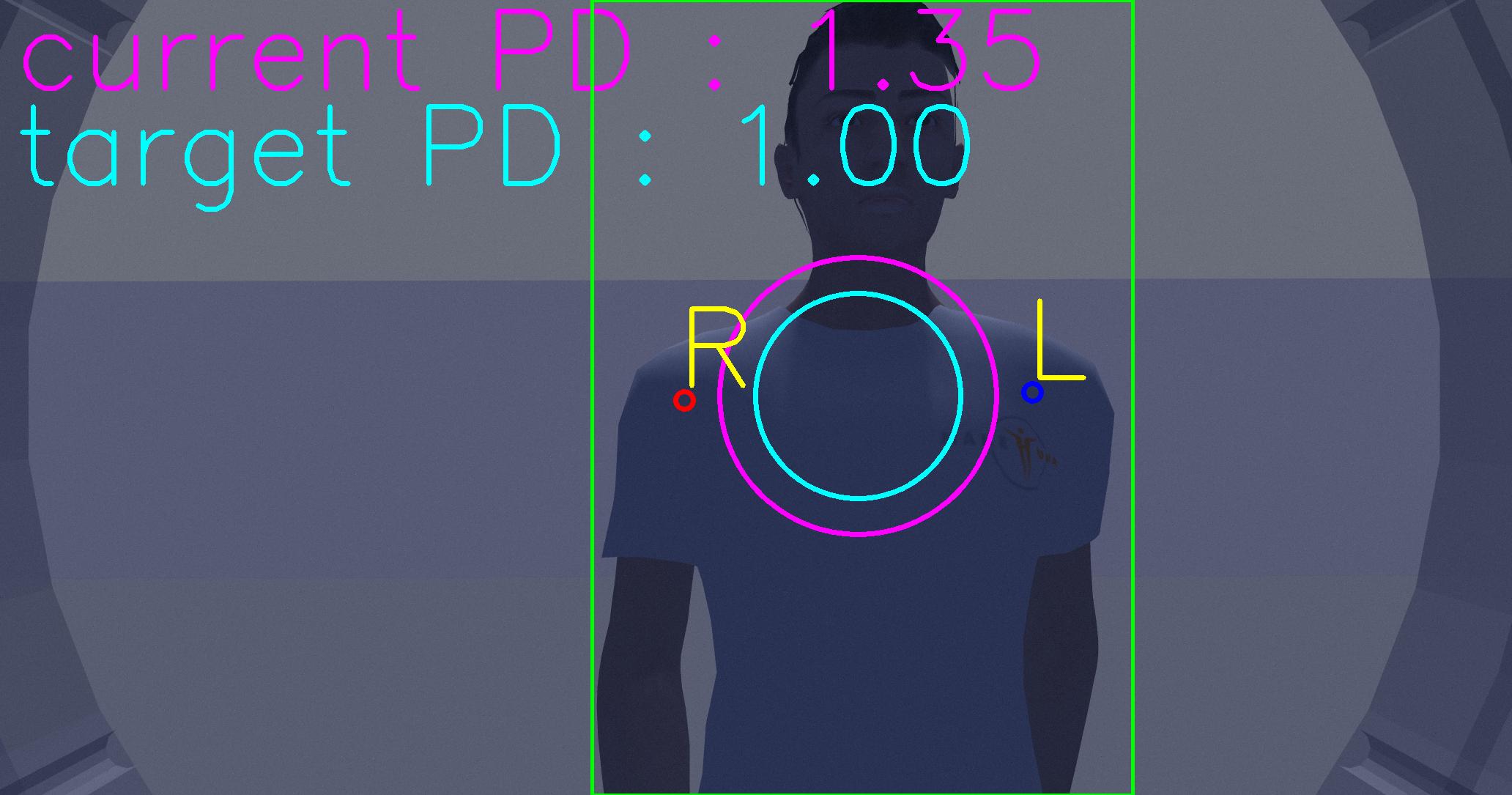}  
      \caption{DRP in Gazebo}
      \label{fig:gazebopd}
    \end{subfigure}
    \begin{subfigure}{.45\textwidth}
      \includegraphics[width=.99\linewidth, trim= 4cm 5cm 0cm 0cm, clip]{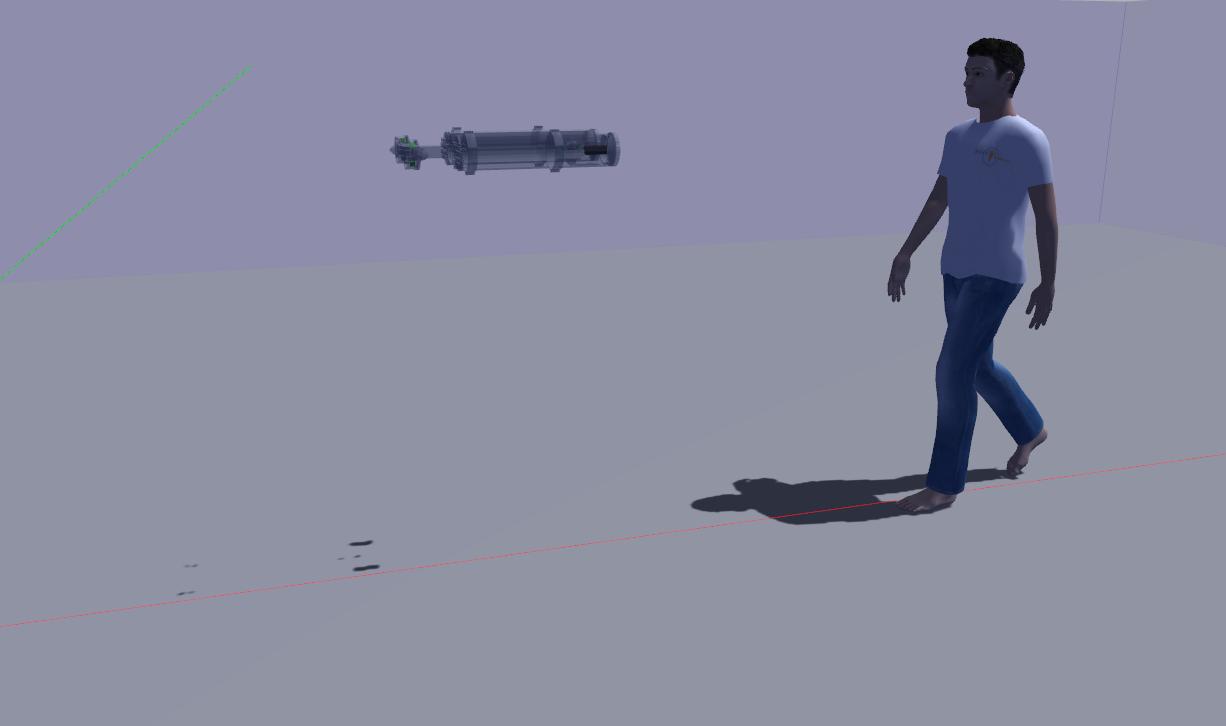}  
      \caption{Third person view in Gazebo}
      \label{fig:personrobot}
    \end{subfigure}
    \begin{subfigure}{.45\textwidth}
      \includegraphics[width=\linewidth, trim= 0cm 3cm 0cm 0cm, clip]{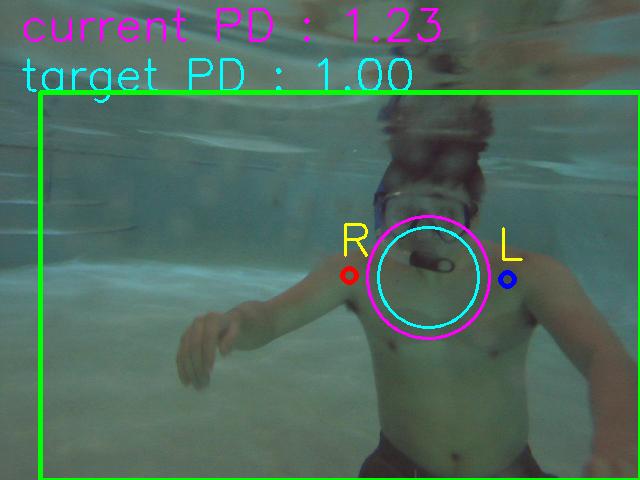}  
      \caption{DRP in pool.}
      \label{fig:personrobot}
    \end{subfigure}
    \begin{subfigure}{.45\textwidth}
      \includegraphics[width=\linewidth, trim= 0cm 0cm 8cm 2cm, clip]{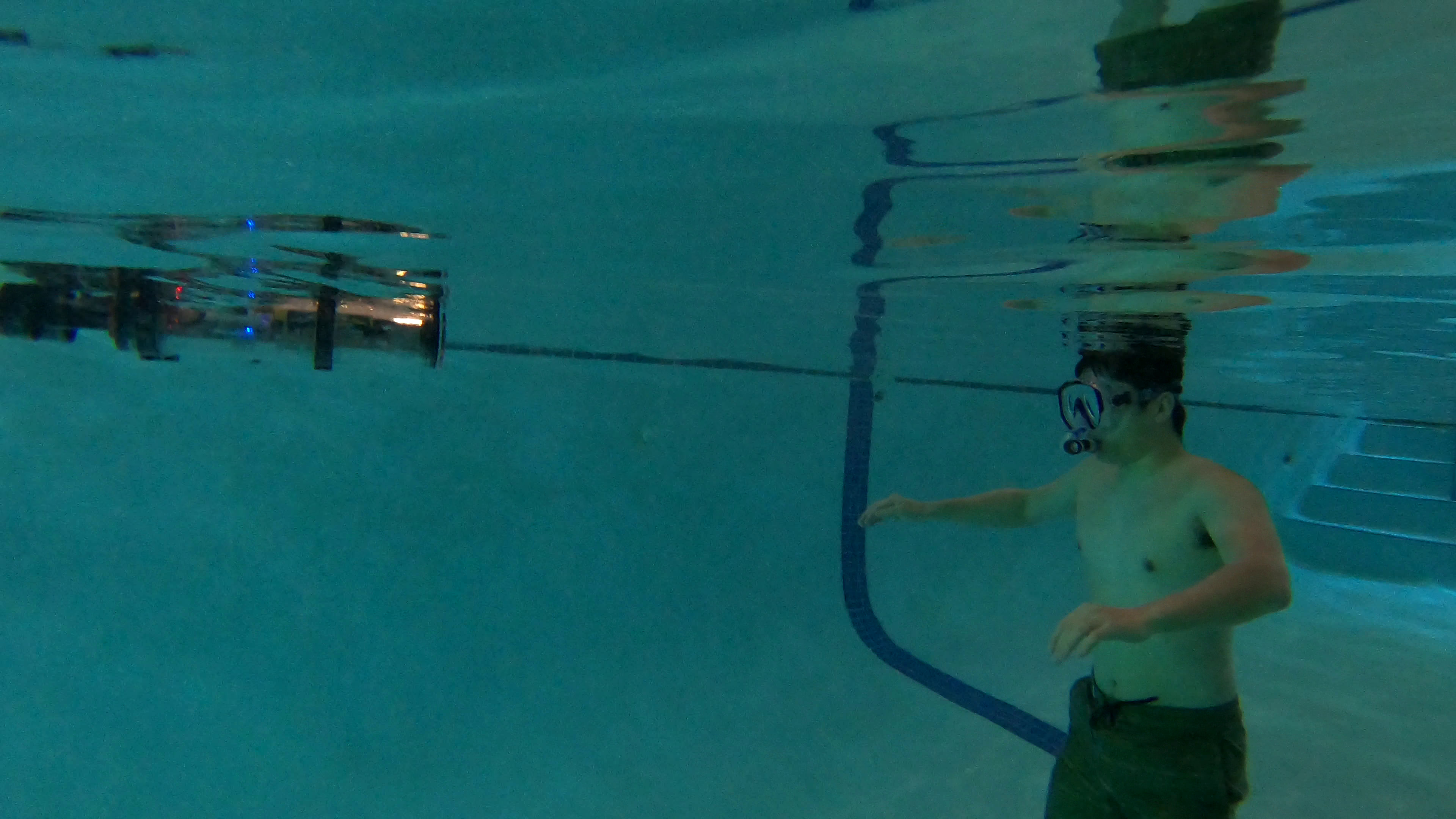}  
      \caption{Third person view in pool.}
      \label{fig:personrobot}
\end{subfigure}
\caption{Diver-relative position (DRP) visualizations with a third person view, displayed in the Gazebo simulator and a pool scene. In the DRP visualization, the center of the circle is the target point while the radius represents pseudo distance.}
\label{fig:gazebo}
\vspace{-4mm}
\end{figure*}

\subsection{Desired Behavior}
\label{sec:aoc:desiderata}
In order to guide our algorithm's development, we draw the following set of desiderata (features desired or required) from our understanding of the AUV interactions we hope to support.
We focus on AUVs working alongside humans in a supporting role, with limited sensing and no global localization, as this is the environment in which we believe our algorithm will be most useful.\\ \\
\noindent\textbf{ADROC Desiderata:}
\begin{itemize}
    \item[\textbf{D1:}] Reliably navigate to a beneficial position for interaction with the diver regardless of the initial location and orientation of both parties.
    \item[\textbf{D2:}] Accurately approach the diver regardless of diver movement after ADROC initiation.
    \item[\textbf{D3:}] Approach the diver in a timely manner.
    \item[\textbf{D4:}] Operate without global localization, 3D diver information, or any off-AUV computation whatsoever. 
\end{itemize} 

These desiderata are integrally tied to our expected deployments and create a very challenging problem. 
As such, we make the following assumptions to simplify the problem allowing us to formulate a feasible solution. \\ \\
\noindent\textbf{ADROC Assumptions}
\begin{itemize}
    \item[\textbf{A1:}] The diver and robot are generally within the visual observation range of one another, and visibility is sufficient to detect the diver in the AUV's cameras.
    \item[\textbf{A2:}] The diver is fully submerged underwater and generally upright with respect to the robot.
    \item[\textbf{A3:}] Only one diver will be present in the scene. 
\end{itemize}

Assumptions \textbf{A1} and  \textbf{A2} are likely to be true in actual deployment scenarios.
While we cannot guarantee ideal visibility conditions in the field, we must assume some level of visibility for the dive to even be occurring. 
\textbf{A2} is necessary to get accurate pose estimations, but could potentially be overcome in the future by retraining a pose estimation algorithm on videos of divers in varied positions.
Lastly, \textbf{A3} is not likely to be true in deployment scenarios, so future versions of this algorithm should account for this by enabling the algorithm to differentiate between divers and approach only a chosen target.
This functionality would require the addition of diver identification capabilities and would be an interesting extension of the ADROC algorithm for future work.
 
\subsection{The ADROC Algorithm}
\label{sec:aoc:algorithm}

ADROC has $3$ components: a \textbf{state machine}, a \textbf{diver-relative pose estimator}, and a Proportional-Integral-Derivative (PID)~\cite{nise_controlsystemsengineering_2015} based \textbf{approach controller} tuned for diver approach operations. 
These components, along with an AUV motion controller and diver perception modules, are pictured on the right side of Fig. \ref{fig:aoc_algorithm} where the interactions between these components may be seen.
The diver perception modules are considered external to the algorithm as they may be replaced with other systems if necessary; these modules are briefly discussed in Section \ref{sec:methods:diver_perception}.
The diver relative position estimator, which is responsible for producing information on the position and distance of a diver relative to the AUV, is addressed in Section \ref{sec:methods:drp}, and we discuss the approach controller in Section \ref{sec:methods:approach}.

Therefore, we largely focus on the state machine portion of ADROC in this section, which is pictured on the left side of Fig. \ref{fig:aoc_algorithm}.
The state machine consists of four states and two conditions which manage transitions between them. 
To begin, the state machine checks if a diver is currently visible in the scene. 
If there is an estimate of the diver's relative position (DRP) available, it transitions from the INIT state to the APPROACH state, which activates the approach controller.
During the APPROACH state, the approach controller visually servos to the target diver, continually checking if a diver remains visible. 
If visual on the diver is lost during the APPROACH state, or if there is no diver visible at the initial state, the AUV transitions to the SEARCH state.
The SEARCH state triggers the AUV's search behavior, which is currently a simple slow yaw motion, turning in circles and scanning for divers.
Once a diver is visible, the SEARCH state transitions to the APPROACH state.
Finally, the APPROACH state is successfully terminated once the diver-relative position is stable at a point close to the ideal position within a configurable margin of error.
Once the AUV is stable in that position, the state machine transitions to the CONCLUDE state, which currently simply terminates the ADROC algorithm.
To understand how the APPROACH state functions and how the conditions responsible for state transition are determined, we must explain the other components of ADROC.

\section{METHODOLOGY}
\label{sec:methods}

\subsection{Diver Perception Modules}
\label{sec:methods:diver_perception}
ADROC is currently designed to operate with diver bounding boxes and diver body pose estimations, which can be obtained from any source. 
We present the modules currently in use, which operate on monocular images, allowing ADROC to be used by any AUV with at least one camera.
The modularity of our approach allows the adaptation of new sensors or perception algorithms in the future.

\subsubsection{Diver Detection}
\label{sec:methods:diver_perception:detection}
While a number of approaches to diver detection have been proposed~\cite{sattar_tracking_2007, islam_2018_toward}, recent work~\cite{delangis2020analysis} profiled the efficacy of popular deep detectors for the diver detection task.
Based on this analysis, we selected a YOLOv4-Tiny diver detector trained on VDD-C~\cite{delangis2020analysis} due to its relatively high accuracy with a fast inference time on embedded hardware. 
The detector outputs a set of detections with associated confidences, from which we select only the highest confidence detection (due to assumption \textbf{A3}).

\subsubsection{Diver Body Pose Estimation}
\label{sec:methods:diver_perception:estimation}
Body pose estimation aims to locate a set of body joints from a given image. 
We obtain a diver pose with a real-time pose estimation implementation~\cite{trt2020} (a TensorRT implementation of human pose estimation~\cite{cao2017realtime, xiao2018simple}). 
From the pose, we only utilized the left and right shoulder joint coordinates of a single body pose detection for our algorithm.

\subsection{Diver-Relative Position Estimation}
\label{sec:methods:drp}
With the diver perception information being provided by our diver detector and pose estimation modules, we process the resultant data to produce a diver-relative position estimate (DRP), comprised of a \textbf{target point} (TP) and \textbf{pseudo distance} (PD). 
This data is used by our approach controller to center the diver in the frame of the camera and to reach a desired distance from the diver.
In calculating both the target point and pseudo distance, we utilize all available information from both the bounding boxes and shoulder coordinates returned by our diver perception modules.
Our algorithm is robust to missing information from either module.

\subsubsection{Target Point}
\label{sec:methods:drp:point}
We define the target point based on bounding box information to be the centroid of the bounding box.
Alternatively, based on body pose estimates, we define the target point as the center point between shoulder joints. 
If both bounding boxes and body pose estimates are available, the target point is defined as the mean of these two points, but if only one is available, it is used as-is.

\subsubsection{Pseudo Distance}
\label{sec:methods:drp:pd}
Our algorithm is designed to work with only monocular vision available so as to achieve our desired functionality even on the least sensor-equipped AUVs.
This creates a challenge, as monocular images do not contain sufficient information to accurately estimate distances, which we need to navigate to an appropriate distance relative to the diver.
Common approaches to estimate the range to a diver would be to use sonar data or calculate disparity and estimate depth from stereo imagery, which has its own challenges, introduced in \ref{sec:related:distance}.
We can overcome this barrier without introducing new sensors and make a general estimate as to the distance between the diver and AUV utilizing the shoulder width of the diver, called \textit{Biacromial breadth}~\cite{mcdowell2009anthropometric}.
Biacromial breadth can differ based on demographic and individual variation, but we use average data available from national surveys~\cite{mcdowell2009anthropometric}, with the option of fine-tuning our estimate with specific measurements of diver shoulder width for an individual.
For body pose estimation input, the shoulder width is obviously the distance between the estimated shoulder points, but for diver detector input we utilize the width of the bounding box for a rough estimate of shoulder size. 
\begin{table}[H]
    \centering
    \caption{The Pseudo Distance (PD) metric based on distance \textit{d} between a diver and robot.}
    \vspace{3mm}
    \begin{tabular}{cc@{}}
     \textbf{Distance} $\mathbf{(d(mm))}$ &\textbf{PD}\\
    \cmidrule[0.9pt](l){1-2}
           \centering$d > D_{ideal}$ & $0 < PD < 1$ \\
           \centering $d=D_{ideal}$ & $PD = 1$ \\
           \centering $d < D_{ideal}$ & $PD > 1 $ \\
           \cmidrule[0.9pt](l){1-2}
    \end{tabular}
        \label{tab:pd}
\end{table}


We propose a metric embedding distance information, \textit{Pseudo Distance (PD)}, using diver detection and pose estimation as individual sources of shoulder width information (Eq~\ref{eq:pd}). 
Psuedo distance is defined as inversely proportional to the ideal interaction distance.
It is greater than $0$ at any distance, less than $1$ when farther from the diver than desired, $1$ at the ideal distance, and greater than $1$ when the robot is closer than the ideal distance (Table~\ref{tab:pd}). 

The following camera sensor, image, and physical specifications are used to develop PD: CMOS sensor size ($mm$), focal length ($mm$), image size ($pixel$), and shoulder width of a diver ($mm$, $pixel$). The detailed steps are:

\begin{enumerate}
    \item Empirically select an ideal distance $D_{ideal} (mm)$ for interaction between a diver and AUV.
    \item Take the average shoulder width reported in \cite{mcdowell2009anthropometric} as the width of the diver's shoulder. Alternatively, measure the true shoulder width of the diver.
    \item Measure the shoulder width in image pixels ($w_{shoulder}$) from an image with the diver at the ideal distance $D_{ideal} (mm)$.
    \item Select $target\_shoulder\_ratio$ as the ratio of the shoulder width ($w_{shoulder}$) to the image width ($w_{img}$).
\end{enumerate}

The $target\_shoulder\_ratio$ is separately defined for diver detection and pose estimation information, as the source of the shoulder width estimate varies drastically in accuracy, with body pose estimation being more accurate. 
This is due to the fact that a bounding box changes size significantly based on the diver's body pose (\eg a diver with open arms yields a much wider bounding box than one with arms held at the side), while the distance between shoulder joints does not change based on the rest of the diver's body pose.
We use $0.3$ for diver detection PD and $0.2$ for pose estimation PD.

\begin{equation}\label{eq:pd}
    PD = \frac{w_{shoulder}}{w_{img} \times target\_shoulder\_ratio}
\end{equation}

Whenever body pose estimation information is available, we default to using a pseudo distance based on pose data, as it is significantly more accurate.
However, when such information is not available, we fall back to our estimate based on diver detection bounding boxes.



\subsection{Approach Controller}
\label{sec:methods:approach}
The calculated DRP is passed on to our approach controller, which manages the approach procedure based on the target point and pseudo distance estimations produced by the DRP estimator.
The approach controller maintains three separate PID controllers for the three axes of control it has: one controller for surge (forward and back), one for yaw (left and right), and one for pitch (up and down). 
These controllers are based off of error measurements (Eq. \ref{eq:fwd_error}-\ref{eq:pitch_error}) comparing the target point to the image's center point and comparing pseudo distance to the ideal pseudo distance, which is defined as 1.0 (see Table \ref{tab:pd}).

\begin{equation}\label{eq:fwd_error}
    error\_forward = 1.0 - PD
\end{equation}
\begin{equation}\label{eq:yaw_error}
    error\_yaw = \frac{target\_x - center\_x}{image\_width}
\end{equation}
\begin{equation}\label{eq:pitch_error}
    error\_pitch = \frac{target\_y - center\_y}{image\_height}
\end{equation}

The PID controllers in the approach controller operate on these errors in the standard fashion~\cite{nise_controlsystemsengineering_2015}, which we will not detail here.
Their parameters are tuned separately based on experimental results.
The approach controller constantly calculates motor control values based on the available target point and pseudo distance, but only applies those controls to the motors (resulting in motion) when enabled by the ADCROC state machine switching to the APPROACH state.
The approach controller limits its speed to $60\%$ of the AUV's maximum velocity for safety.

\input{fig/pool}
\section{Experimentation}
\label{sec:experiments}

\subsection{Experimental Platforms}
\label{sec:experiments:platforms}
The AUV we used as our test platform for this work was the LoCO AUV~\cite{loco_paper_2020}, which is a modular, low-cost, open-source AUV equipped with dual monocular cameras and three thrusters. 
All the computation required for ADROC was done onboard, primarily using a Nvidia Jetson TX2 mobile GPU. 
Additionally, we used a LoCO simulation in ROS Gazebo~\cite{koenig_gazebo_2004} (Fig. \ref{fig:gazebo}) as a tool for developing the algorithms and evaluating them, particularly, the pseudo distance (Eq~\ref{eq:pd}) and the approach controller. 

\begin{figure*}[ht]
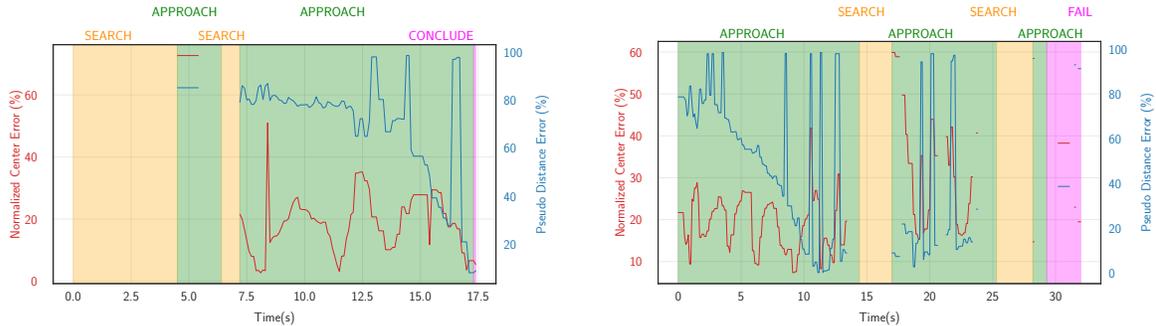

     \centering
     \vspace{2mm}
     \begin{subfigure}[b]{0.48\linewidth}
         \centering
         \scalebox{0.2}{\input{fig/jiyeon_success_big.pgf}}
         \caption{Success example with narrow shoulder width diver ($35cm$), long distance ($9m$), and turned-away angle($90^{\circ}$)}
         \label{fig:success}
     \end{subfigure}
     \begin{subfigure}[b]{0.48\linewidth}
         \centering
         \scalebox{0.2}{\input{fig/mike_fail_big.pgf}}
         \caption{Failure example with wide shoulder width diver ($45cm$), long distance ($9m$), and head-on angle($0^{\circ}$)}
         \label{fig:fail}
     \end{subfigure}
        \caption{(a) LoCO was turned away from a diver at the beginning. Through the SEARCH and APPROACH states, LoCO detected the diver and placed itself at the designed distance from the diver, (b) LoCO was facing a diver and started with the APPROACH state. However, spikes in PD error (bounding box only DRP estimation) caused unstable control and failure.}
        \label{fig:errors}
        \vspace{-4mm}
\end{figure*}

\begin{table}[h]
    \caption{The success rates and average operation time of ADROC based on the trial, distances, and angles.}
    \centering
    \vspace{3mm}
    \begin{tabular}{lcc@{}}
    \textbf{Experiment} &\textbf{Success} &\textbf{Avg. Time}\\
    \cmidrule[0.9pt](l){1-3}
    Exp. \#1 (clear pool) & 88.9\% & 25s \\
    Exp. \#2 (cloudy pool) & 66.7\%&28s \\
    & & \\
    \textbf{Distance} &\textbf{Success} &\textbf{Avg. Time}\\
    \cmidrule[0.9pt](l){1-3}
    3 meters & 92.6\%&21s \\
    6 meters & 83.3\%&28s \\
    9 meters & 68.5\%&29s \\
        & & \\
    \textbf{Angle} &\textbf{Success} &\textbf{Avg. Time}\\
    \cmidrule[0.9pt](l){1-3}
    0\degree & 77.8\%&22s \\
    45\degree & 83.3\%&24s \\
    90\degree & 83.3\%&33s \\
    \cmidrule[0.9pt](l){1-3}
    \textbf{Overall} &\textbf{81.5\%} &\textbf{26s}\\
    
    \end{tabular}
    \label{tab:aggregate_data}
\end{table}

\subsection{Pool Experiments}
\label{sec:experiments:pool}
A set of pool experiments was performed to validate the ADROC algorithm's ability to successfully approach a diver.
These experiments took place in two different pools, one with clear water and bright, even lighting, and another with murky water and dim, irregular lighting.
A total of $9$ divers were used as the target of the approach algorithm, with shoulder widths between $33cm$ and $48cm$.
Six of the divers were tested in the clear pool (Experiment \#1) and three in the cloudy pool (Experiment \#2).
In each trial, the AUV was made to approach the diver from the combination of three initial distances ($3$ meters, $6$ meters, $9$ meters) and from three initial angles between the AUV and the diver ($0$\degree, $45$\degree, $90$\degree).
These variables of distance and orientation combine to create nine distinct conditions of AUV approach. 
For each condition, two trials were conducted per diver for a total of $162$ trials.
For each trial, we allowed the AUV to enter the SEARCH state up to twice after the first APPROACH state.
If the AUV reached the CONCLUDE state at the desired position before entering the third SEARCH state, we considered the cases as a success, but as soon as the AUV entered the third SEARCH state after an APPROACH, the case was deemed a failure. 
Additionally, we recorded the time taken for each case, regardless of success.

\subsection{Results}

\subsubsection{Types of Failures}
Of the $162$ recorded trials, $30$ failures were recorded.
These failures can be grouped into three categories: \textbf{search failures}, \textbf{early conclusion}, and \textbf{no conclusion}.
\textbf{Search failures} ($12$ cases) were characterized by repeated returns to the SEARCH state, usually due to one of two issues: overshooting the initial rotation required for an approach, or receiving inaccurate detection data from the diver perception modules.
Re-tuning the approach controller may help to reduce these failures. 
\textbf{Early conclusion failures} ($11$ cases) were caused by ADROC entering the CONCLUDE state prior to reaching the appropriate relative distance to the diver. 
These occurred almost exclusively for participants with small shoulder widths and were caused by the pseudo-distance based on bounding boxes reaching a stable point at a distance of $5$ meters, prior to the AUV entering the detection range of the diver pose estimation.
This type of failure can likely be entirely resolved by tuning the \textit{target\_shoulder\_ratio} as described in Section \ref{sec:methods:drp:pd}.
Lastly, \textbf{no conclusion failures} ($7$ cases) were caused by ADROC failing to detect a stable position while in the appropriate relative position to the diver.
This failure is likely due to tuning issues with the approach controller, but should also be resolved by further improvement of the approach controller.

\subsubsection{Aggregate Data}
The summarized data in Table~\ref{tab:aggregate_data} represents the averages of our results.
We note that Experiment \#$1$ has a higher overall success rate ($88.9\%$) compared to Experiment \#2, likely due to the better visual environment which led to higher quality perceptual inputs.
Success rates reduce as the initial distance increases, most likely due to reduced accuracy of diver detection and pose estimation, which leads to fewer successful searches.
Average times for an approach increase as well, but this is mostly due to the increased distance that the robot has to travel. 
When considering the effect of the initial angle, we see that approaches from the $0$\degree{} condition have the lowest accuracy, but not by a statistically significant amount. 
This is likely due to random chance, as the difference in success between the $0$\degree{} and the other angle conditions (which have the same success rate of $83.3\%$) is only $3$ failures.

\subsubsection{ADROC Runtime}
All components of ADROC were run on the onboard computers of the LoCO-AUV: the Raspberry Pi $4$ and the Nvidia Jetson TX2.
The diver detector and diver pose estimator, running concurrently on the TX2, ran at $15$ $fps$ and $10$ $fps$ respectively. 
Also on the TX2, the diver relative position estimator ran at a set frequency of $20$ Hz, while the ADROC state machine ran at $10$ Hz. 
The only component which ran on the Raspberry Pi was the approach controller, which runs at a set frequency of $10$ Hz. 
This frequency of the overall system (about $10$ Hz) was sufficient to operate the AUV at relatively low speeds, although an improved frequency is possible with a more powerful GPU.

\subsubsection{Individual Cases}
Fig.~\ref{fig:errors} presents one of each success and failure cases from our experiments. 
In Fig.~\ref{fig:success}, the AUV was at the turned-away angle, thus starting with the SEARCH state. 
During its second APPROACH state (t=$7.3$-$17.3s$), the AUV occasionally failed to detect pose estimates of the participant, and the diver detection estimate was used to yield the PD.
The spikes (t=$12.9s$, $14.4s$, and $16.4s$)in the Pseudo Distance Error were caused when the AUV controlled its distance to the participant based on the diver detection as opposed to body pose estimation.
In Fig.~\ref{fig:fail}, the AUV started with the APPROACH state since it was at the head-on angle and was able to see the participant right away. 
The participant was detected by the diver detection consistently while the pose estimation only failed occasionally.
More frequent PD estimation based on the diver detection caused an overshoot from the approach controller (more spikes from the beginning), and it resulted in failure.
Additionally, the missing data points during the second APPROACH phase (t=$20.7$-$21.2s$) are consistent with system lag, possibly due to processing issues with the camera or some competition between processes for computational resources.

\begin{figure}[t]
\vspace{2mm}
\centering
    \begin{subfigure}{.45\textwidth}
      \includegraphics[width=\linewidth, trim= 21cm 0cm 3cm 0cm, clip]{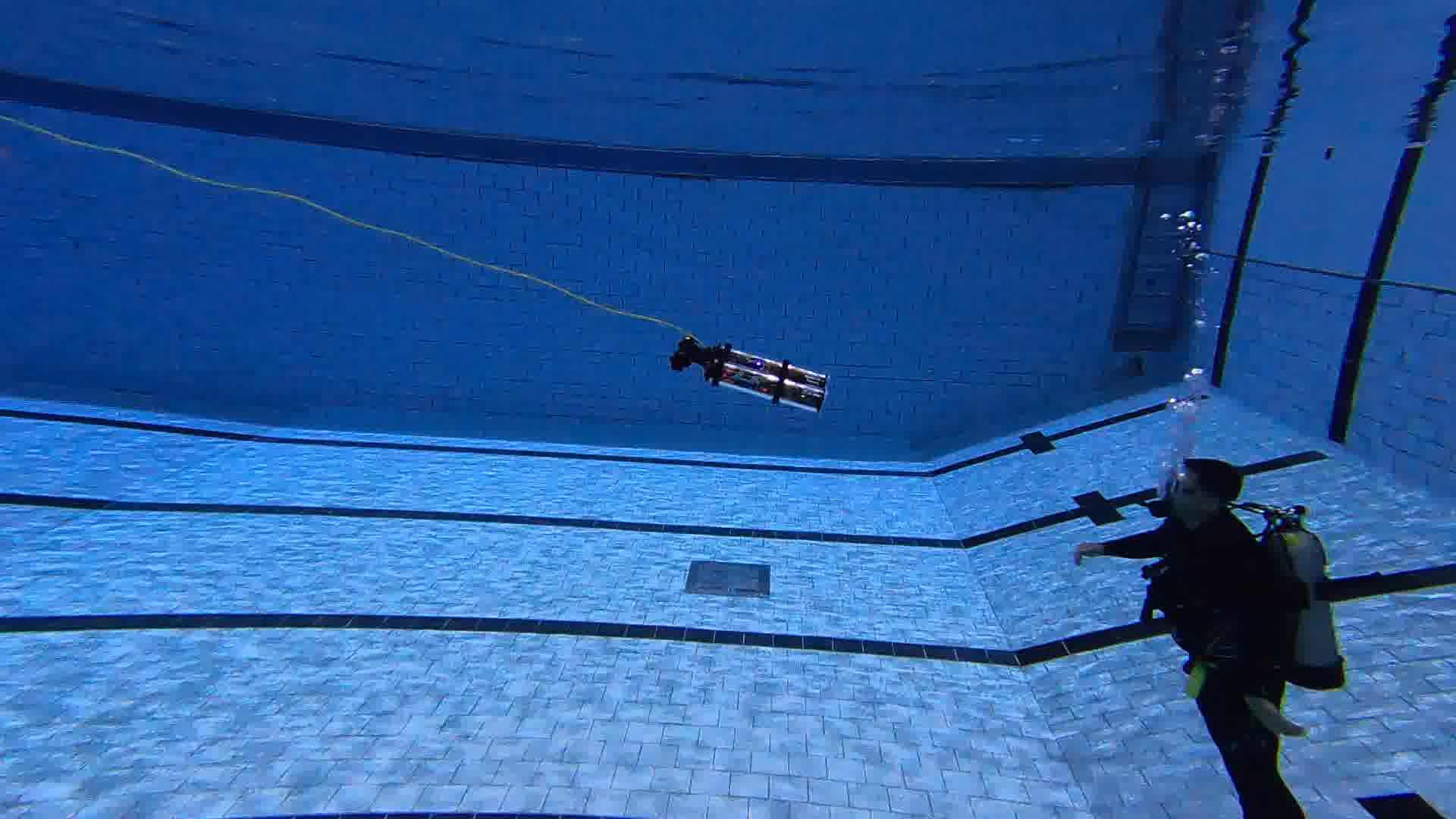}  
      \caption{DRP in pool.}
      \label{fig:personrobot}
    \end{subfigure}
    \begin{subfigure}{.45\textwidth}
      \includegraphics[width=\linewidth, trim= 0cm 0cm 3cm 0cm, clip]{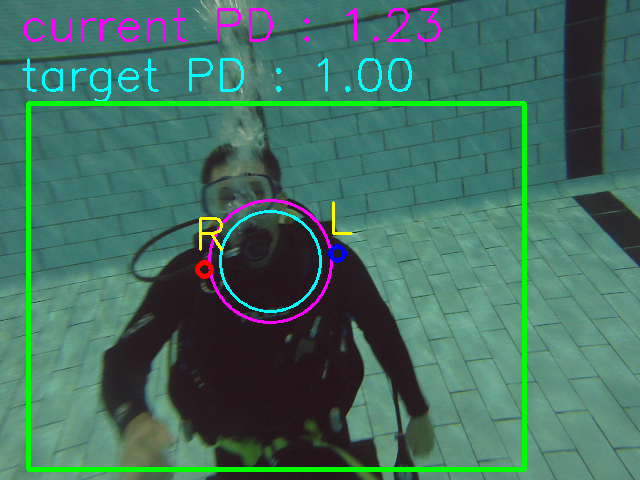}  
      \caption{Third person view in pool.}
      \label{fig:personrobot}
\end{subfigure}
\caption{Diver-relative position (DRP) visualizations for a scuba diver in a deep pool.}
\label{fig:suba}
\vspace{-6mm}
\end{figure}

\subsection{ADROC Limitation Experiments}
In order to explore the limits of ADORC, a small number of trials were performed with more challenging conditions, in a deep pool similar to the pool used for Trial \#$1$.
These trials are not included in the aggregate results above, because of the small number of participants for each.

\subsubsection{Scuba Gear}
We repeated ADROC trials for one participant who was included in the normal trials in scuba equipment such as a wetsuit instead of the general swimwear that our participants wore. 
No reduction in accuracy compared to non-scuba trials was detected.

\subsubsection{Adversarial Conditions}
Trials were also conducted with an adversarial condition, in which the AUV was initially in a $0$\degree{} orientation to the diver, then was pushed $20$\degree{}- $40$\degree{} to the left by study staff. 
This push, meant to mimic a wave knocking the robot away from its chosen path, did not cause a significant reduction in accuracy as the robot was able to reacquire its path.

\subsubsection{Long-Range Condition}
ADROC trials have been conducted up to $15$ meters away from the diver. 
While the algorithm generally still functions, the AUV struggles to effectively switch from SEARCH to APPROACH as the approach controller sends yaw commands of a high intensity causing the AUV to quickly overshoot and miss its target. 
This could likely be improved with pseudo distance-relative tuning of the approach controller.
Currently the effective range of ADROC is limited to approximately $15$m, which is further than the visibility in many open water environments and comparable to the effective range of stereo-based distance estimation.
\section{CONCLUSION}
\label{sec:conclusion}
We present a novel algorithm, Autonomous Diver-Relative Operator Configuration (ADROC), to facilitate diver-AUV interaction for underwater HRI.
In our experiments, ADROC enabled an AUV to robustly approach participants with various shoulder widths from different distances and angles. 
Increased initial distance from the diver caused a reduction in success rates, but no initial angles caused clear differences in success rates.
Additional trials provided an indication that ADROC continues to operate well with targets in full scuba gear and when faced with adversarial conditions, but that the algorithm's success degrades significantly at distances greater than $15$ meters.
The advantages of our method are that it only requires a monocular camera, needs no extra sensors to estimate the distance to a diver, functions without global localization, and runs at real-time speed using on-board hardware.

We plan to extend the ADROC algorithm on multiple fronts. 
First, we will develop a more robust DRP estimate by feeding the raw target point and pseudo distance values to Bayesian filters. 
This will help reduce the negative effect of noisy input data from the various diver perception modules.
Additionally, we plan to reduce the overshooting behavior of the approach controller by continuing the tuning of the controller. 
As previously mentioned, we also plan to expand the algorithm by implementing a variety of search methods to improve the likelihood of successfully locating a diver for approach.
In the same vein, an expansion to the search and approach behaviors of ADROC to enable operation with multiple divers in the scene will allow us to reduce the assumptions we make and increase the applicability of ADROC to varied deployments.
Lastly, we will test ADROC in ocean environments to see how it is affected by the current and visibility issues that are often present in field environments.

Our presented work clearly demonstrates the efficacy, reliability, and potential of our algorithm, despite the minimal information and sensors utilized. 
As it improves, ADROC and similar algorithms will become a standard capability for AUVs, allowing natural and straightforward cooperative work between AUVs and humans underwater.
\vspace{-2mm}

\bibliographystyle{ieeetr}
\bibliography{ref.bib}
\end{document}